\documentclass[runningheads]{llncs}

\usepackage[T1]{fontenc}
\usepackage{graphicx,verbatim}
\usepackage{amsmath}
\usepackage{amssymb}
\usepackage{amsfonts}
\usepackage{bm}
\usepackage{array}
\usepackage[table]{xcolor}
\usepackage{float}
\usepackage{marvosym}

\begin{document}

\title{LINDA: Low Resource Foot Ulcer Segmentation using Local Similarity and Iterative Neural Dynamics}
\titlerunning{LINDA}

\author{Trishaani Acharjee\inst{1,2}\thanks{Equal contribution.} \and
Sudip Chakrabarty\inst{1,2}$^{\star}$ \and
Rajdeep Chatterjee\inst{1,2}$^{\star}$(\Letter) \and
Tathagata Bandyopadhyay\inst{3}}

\authorrunning{T. Acharjee et al.}

\institute{School of Computer Engineering, Kalinga Institute of Industrial Technology, Bhubaneswar, India\\
\email{cse.rajdeep@gmail.com} \and
AmygdalaAI-India Lab, Bhubaneswar, India \and
Department of Informatics, Technical University of Munich, Munich, Germany}

\maketitle
\begin{abstract}
Foot ulcers are a leading cause of lower limb amputation, and the highest burden often falls on low and middle income regions where specialists and computing hardware are both scarce. Wound monitoring tools are most useful there when they are accurate yet small enough to run on ordinary devices, but accuracy usually comes from heavy networks. We present LINDA, a compact model that grounds segmentation in an explicit structural prior instead of raw intensity. A Local Self-Similarity (LSS) map, computed directly from the image and stable under local lighting changes, is fused into a ResNet-34 encoder, and a small Liquid Time-Constant module refines the boundary over four iterative steps. With every model trained under one protocol on the FUSeg 2021 dataset, LINDA obtains the best Dice (86.68\%), IoU (79.67\%), and boundary accuracy (HD95 6.37 pixels) in our comparison while using only 25.70M parameters, fewer than the U-Net and transformer baselines. As the LSS map is a fixed function of the input, it exposes the structural cues the model follows in a readable form. Together this makes LINDA an accurate, inspectable, and lightweight option for wound assessment in resource-constrained care.


\end{abstract}

\section{Introduction}
Foot ulcers are a primary cause of non-traumatic lower-limb amputation \cite{1_armstrong2017diabetic}, and precise delineation of the wound bed is needed to track healing and plan treatment \cite{2_cassidy2021dfuc2020}. Manual planimetry, the clinical gold standard, is subjective, slow, and varies between observers, so automated segmentation has become a priority. In many settings, and especially in low and middle income regions, this must run with limited specialist time and limited computing hardware, which favors models that are accurate yet small enough for modest devices.

Despite recent advances \cite{alkhalefah2025advancing}, automated foot ulcer segmentation remains a hard problem. Ulcers vary widely in scale, lighting, and texture. The transition zones between necrotic tissue, granulation tissue, and healthy skin are often ambiguous, with low contrast and diffuse gradients \cite{3_goyal2020recognition}. Widely used architectures such as U-Net variants \cite{4_ronneberger2015u} and Vision Transformers (ViTs) \cite{5_chen2021transunet} perform well on many medical segmentation tasks, yet wound boundaries can remain imprecise when ulcer shapes are irregular and contrast with healthy skin is low. Convolutional encoders have a limited receptive field, while the patch tokenization in ViTs can blur fine boundary detail and lead to oversmoothed masks \cite{6_hatamizadeh2022unetr}. Most of these models also predict the mask in a single forward pass, unlike the way a human reader traces a boundary, revisiting and adjusting it as local and global cues are weighed.

To address these limitations, we propose \textbf{LINDA}, a boundary aware architecture for foot ulcer segmentation. First, a Local Self-Similarity (LSS) map is computed directly from the input image and fused into a pretrained ResNet-34 backbone, giving the early layers explicit local texture awareness. Second, a Liquid Time-Constant (LTC) network \cite{7_hasani2021liquid} at the bottleneck acts as an iterative refinement loop that resolves boundary ambiguities. Finally, the structural prior stays visible: because the LSS map is computed before encoding and used as the target of the boundary loss, it offers a concrete, readable view of the cues the network follows, in the spirit of explainable AI in clinical imaging \cite{11205342,akman2024audio,van2022explainable}.

Our contributions are as follows. (1) An additive LSS fusion module injects local patch-correlation statistics into the early encoder features, anchoring the network to tissue boundaries without disturbing the pretrained weights. (2) An LTC-driven refinement loop at the bottleneck uses iterative dynamics to refine a global spatial token that conditions the decoder. (3) A Boundary Alignment Loss ties the prediction to the LSS map, giving the lowest boundary error in our comparison together with strong overlap accuracy at a low parameter count. Overall, these components form a compact, interpretable foot ulcer segmentation model that matches the strongest baselines on region overlap and clearly improves boundary precision, while using far fewer parameters, a good fit for resource-constrained clinical settings.

\vspace{-0.3cm}
\section{Previous Work}
Automated wound segmentation is built mostly on the U-Net family \cite{4_ronneberger2015u} and, more recently, on transformer encoders \cite{5_chen2021transunet,6_hatamizadeh2022unetr}. Several methods target wounds directly. Wang et al. \cite{wang2020fully} trained fully convolutional networks for wound region extraction, and WSNet \cite{oota2023wsnet} added a wound-oriented pipeline with dedicated pre-processing. On the FUSeg 2021 benchmark, FUSegNet \cite{dhar2024fusegnet} reported the strongest published result and remains the reference point for this dataset. State-space (Mamba) models have also entered the field, from lightweight ulcer designs like DFU-MambaLiteUNet \cite{nataliani2025dfu} to general backbones such as MambaVision \cite{hatamizadeh2025mambavision} and MedMamba \cite{yue2024medmamba}, which we include as baselines. A recent benchmark, WoundAmbit \cite{borst2025woundambit}, finds that general-purpose segmentation models can match or beat task-specific designs once training and evaluation are standardized, and it argues for attention to out-of-distribution generalization, efficiency, and interpretability. This work takes that view as its starting point: rather than competing on overlap alone, it targets tight boundaries, a small parameter budget, and a structural prior a reader can check. Evaluation on further collections such as Medetec \cite{thomas2017medetec} is left for future work.

Local self-similarity was introduced by Shechtman and Irani \cite{shechtman2007matching} to match patterns across images and videos through internal patch correlations rather than raw intensity. We reuse this idea as a deterministic prior and fold it into the encoder. For interpretability, post-hoc attribution such as Grad-CAM and SHAP is widely used in medical imaging \cite{van2022explainable,bhalla2023discriminative}, but these maps are produced after training and can be unstable. Our LSS maps are fixed functions of the input and serve as the target of the boundary loss, so they give a stable, readable view of the structural evidence the model is trained to follow.
\vspace{-0.3cm}
\section{Methodology}

\begin{figure}[t]
\centering
\includegraphics[width=11cm, height=5.3cm]{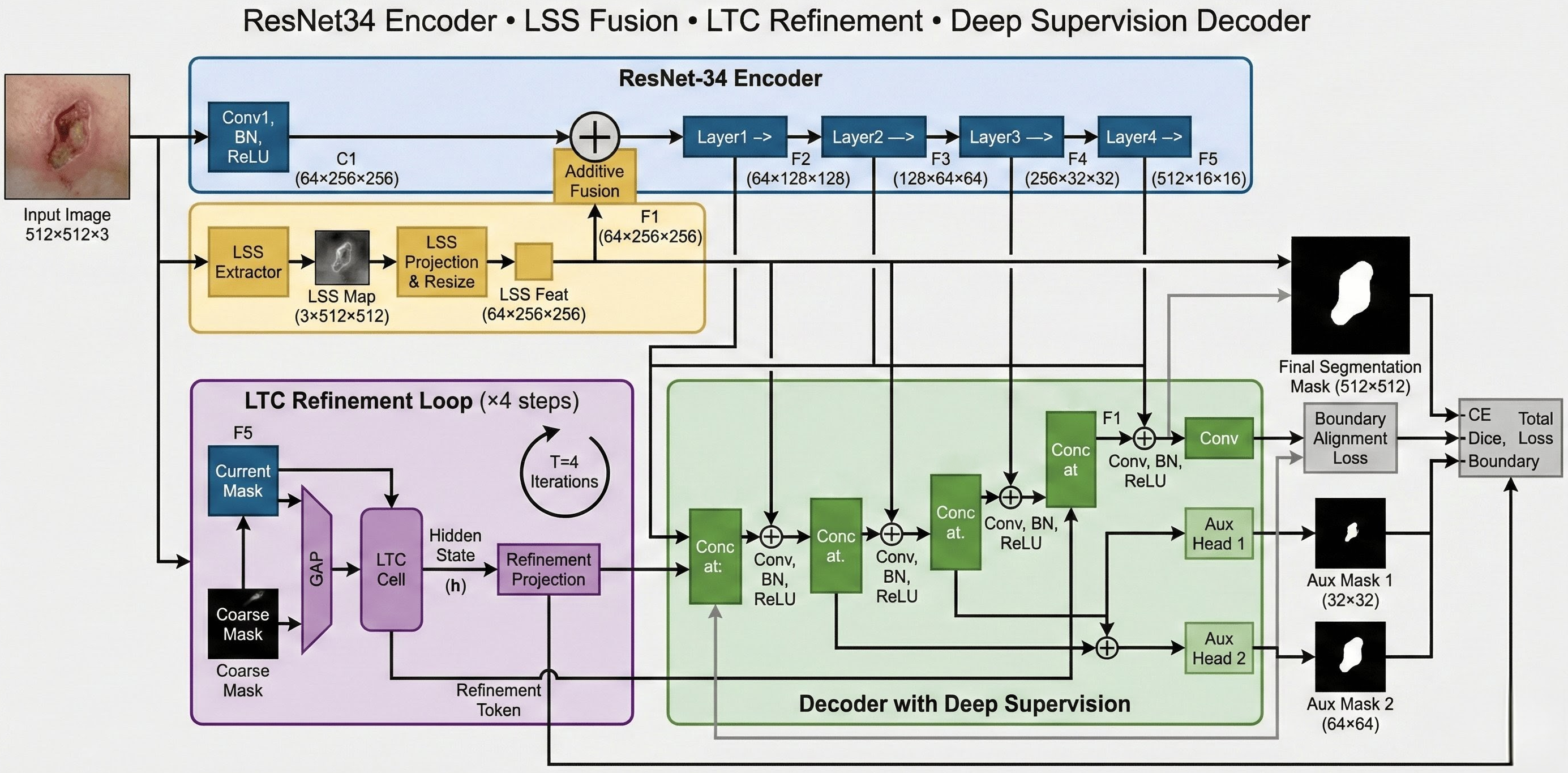}
\caption{The proposed LINDA, illustrating the integration of the ResNet-34 encoder, Additive LSS Fusion, LTC Refinement Loop, and Deep Supervision Decoder.}
\label{fig1}
\end{figure}

LINDA is an end-to-end, boundary-aware segmentation network for clinical wound images; Figure~\ref{fig1} gives an overview. It has four modules: a ResNet-34 encoder, an additive LSS fusion block, a Liquid Time-Constant (LTC) refinement loop, and a deep-supervision decoder. Different modules are discussed in the following subsections.

\vspace{-0.3cm}
\subsection{Local Self-Similarity (LSS) Fusion}
To add an explicit structural prior, the LSS Extractor operates on the input image $\mathbf{I}$ before any deep encoding. Densely overlapping local patches $\mathbf{p}_i \in \mathbb{R}^{K \times K \times 3}$ are extracted, and to stay robust to the uneven illumination common in wound photography, each is zero-centered and normalized (Eq.~\ref{eq:normalization}):
\begin{equation}
\label{eq:normalization}
\hat{\mathbf{p}}_i = \frac{\mathbf{p}_i - \mu(\mathbf{p}_i)}{\|\mathbf{p}_i - \mu(\mathbf{p}_i)\|_2 + \epsilon}
\end{equation}
where $\mu(\mathbf{p}_i)$ is the patch mean and $\epsilon$ is a small stability constant. Since the patches are unit-normalized, the dot product between a central patch $\hat{\mathbf{p}}_c$ and each of its $N$ neighbors within radius $R$ (Eq.~\ref{eq:similarity}) equals their cosine similarity, which compares patch structure independently of local brightness and contrast:
\begin{equation}
\label{eq:similarity}
S(\hat{\mathbf{p}}_c, \hat{\mathbf{p}}_n) = \hat{\mathbf{p}}_c \cdot \hat{\mathbf{p}}_n
\end{equation}
This reuses the local self-similarity descriptor of Shechtman and Irani \cite{shechtman2007matching} as a dense, deterministic prior. Aggregating the scores at each location gives a 3-channel LSS map $\mathbf{M}_{LSS} \in \mathbb{R}^{3 \times 512 \times 512}$, holding the mean ($\mu$), maximum ($\max$), and standard deviation ($\sigma$) of the local correlations. A $3 \times 3$ convolution with bilinear interpolation projects it to $\mathbf{F}_{LSS} \in \mathbb{R}^{64 \times 256 \times 256}$, which is added to the first ResNet feature map $\mathbf{C1}$ to form the boundary-conditioned map $\mathbf{F1} = \mathbf{C1} \oplus \mathbf{F}_{LSS}$ that guides the remaining encoder layers.

\vspace{-0.3cm}
\subsection{Latent Refinement (LTC)}
To capture global dependencies, the bottleneck LTC loop ingests the high-level feature map $\mathbf{F5} \in \mathbb{R}^{512 \times 16 \times 16}$ together with a dual-mask input, the initial coarse mask and the evolving current mask. Features and masks are globally average-pooled (GAP) and concatenated into a 514-D vector $\mathbf{x} = [\text{GAP}(\mathbf{F5}); \allowbreak \text{GAP}(\mathbf{Y}_{masks})]$. The hidden state $\mathbf{h}$ then evolves by the ODE of Eq.~\ref{eq:ltc_ode}:
\begin{equation}
\label{eq:ltc_ode}
\frac{d\mathbf{h}}{dt} = \left[ -\mathbf{h} + \sigma_{rel}(W_h \mathbf{h} + W_{in} \mathbf{x} + b) \right] \odot \frac{1}{\tau(\mathbf{x})}
\end{equation}
where $\sigma_{rel}$ is ReLU and $\tau(\mathbf{x}) = \text{softplus}(W_{\tau} \mathbf{x}) + \epsilon$ is an input-dependent time constant that modulates the cell's memory. The system is solved by Euler discretization ($\mathbf{h}_{t+1} = \mathbf{h}_t + \Delta t\, \tfrac{d\mathbf{h}}{dt}$) over $T=4$ steps, and the final state is projected into a Refinement Token, a single global vector that conditions the decoder and steers the predicted boundary.
\vspace{-0.3cm}

\subsection{Optimization and Deep Supervision}
The network is trained with multi-scale deep supervision. As shown in Fig.~\ref{fig1}, it produces the final mask $\mathbf{\hat{Y}}$ and two auxiliary predictions, $\mathbf{Aux}_1 \in \mathbb{R}^{32 \times 32}$ and $\mathbf{Aux}_2 \in \mathbb{R}^{64 \times 64}$. The total objective combines Binary Cross-Entropy (BCE), Dice, and a geometric Boundary Alignment Loss (BAL):
\begin{equation}
\label{eq:total_loss}
\mathcal{L}_{total} = \sum_{k \in \{m, a1, a2\}} \lambda_k \left( \mathcal{L}_{BCE}^{(k)} + \mathcal{L}_{Dice}^{(k)} \right) + \lambda_b \mathcal{L}_{BAL}
\end{equation}
with weights $\lambda_{m}=1.0, \lambda_{a1}=0.4, \lambda_{a2}=0.2$, and $\lambda_{b}=0.5$. The BAL grounds the prediction in tissue structure by matching the Sobel gradients $\nabla(\cdot)$ of the predicted probability map to those of the deterministic LSS Mean channel $\mathbf{M}_{LSS}^{\mu}$:
\begin{equation}
\label{eq:bal_loss}
\mathcal{L}_{BAL} = \text{MSE}(\nabla(\sigma(\mathbf{\hat{Y}})), \nabla(\mathbf{M}_{LSS}^{\mu}))
\end{equation}
so it acts as a structural regularizer that penalizes predictions departing from the local tissue transitions the LSS extractor marks.

\vspace{-0.3cm}
\subsection{Interpretable Structural Priors}
Because the LSS extractor is deterministic, its three channels carry a fixed, readable meaning rather than the shifting saliency of post-hoc attribution \cite{bhalla2023discriminative}. \emph{LSS Mean ($\mu$)} measures local homogeneity and highlights uniform regions such as granulation tissue in the wound bed. \emph{LSS Max} keeps the strongest local correlation in each neighborhood, helping thin structures survive downsampling. \emph{LSS Std ($\sigma$)} responds to high local variance and marks the sharp transition between wound and healthy skin. These maps are not a claim about what individual network weights compute. They are deterministic inputs to the encoder, and the Mean channel is also the target of the boundary loss (Eq.~\ref{eq:bal_loss}), so this part of the pipeline can be inspected directly.

\vspace{-0.3cm}
\subsection{Compute, Data and Inference Time}
For both training and evaluation, we have used single NVIDIA A100 GPU with 40 GB VRAM. We have performed all the experiments on MICCAI Foot Ulcer Segmentation (FUSeg) Challenge dataset \cite{wang2024fuseg}, details of which are described in sec \ref{data}. Our model takes roughly $5$ hours to train and $\sim45$ ms per image, on average, for inference.
\vspace{-0.3cm}
\section{Results}

\subsection{Dataset and Implementation Details}\label{data}
\begin{figure}[!htb]
\centering
\includegraphics[width=11cm, height=4.7cm]{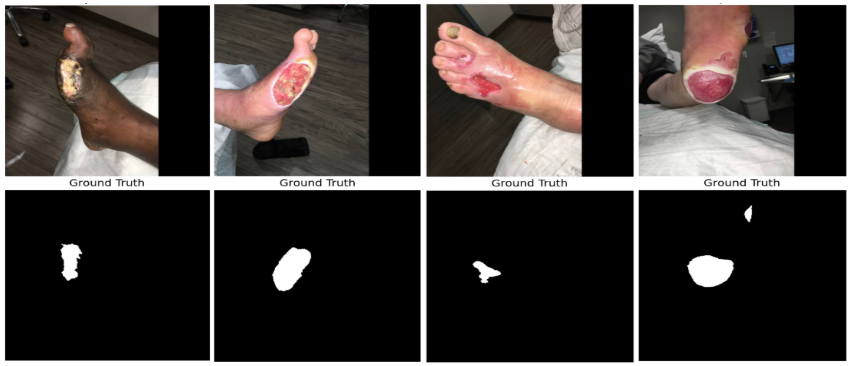}
\caption{Sample images and ground-truth masks from the FUSeg 2021 challenge dataset.}
\label{fig2}
\end{figure}

We use the MICCAI Foot Ulcer Segmentation (FUSeg) Challenge dataset \cite{wang2024fuseg}, which has 1{,}210 images (810 train, 200 validation, 200 test) from 889 patients under varied lighting and backgrounds. The test labels are held private by the organizers, so all metrics are reported on the official validation set. Sample images are shown in Figure~\ref{fig2}.
Images and masks are processed at $512 \times 512$, the grid also used for prediction. LINDA is implemented in PyTorch 2.5.1 and trained on an NVIDIA A100. For a fair comparison, every model uses the same setup: 100 epochs, batch size 2, the Adam optimizer, and the checkpoint with the best validation Dice.

The ResNet-34 encoder uses ImageNet-pretrained weights, while the LSS Extractor and LTC loop are trained from scratch. The task is binary, wound against background; the necrotic and granulation regions noted earlier describe the visual difficulty inside the wound, not separate classes. We report Dice (DSC), IoU, and the 95th-percentile Hausdorff Distance (HD95), all on the wound class.
\vspace{-0.35cm}

\begin{table}[!ht]
\centering
\caption{Quantitative comparison of LINDA against representative segmentation methods under a unified training protocol.}
\label{tab:sota_comparison}
\renewcommand{\arraystretch}{1.2}
\begin{tabular}{ >{\arraybackslash}m{3.6cm} | >{\centering\arraybackslash}m{1.5cm} >{\centering\arraybackslash}m{1.5cm} | >{\centering\arraybackslash}m{1.5cm} >{\centering\arraybackslash}m{1.5cm} >{\centering\arraybackslash}m{1.5cm} }
\hline
\textbf{Method} & \textbf{Params (M)} & \textbf{FLOPs (G)} & \textbf{Dice (\%)} ($\uparrow$) & \textbf{IoU (\%)} ($\uparrow$) & \textbf{HD95} ($\downarrow$) \\
\hline
VGG16-UNet \cite{simonyan2015very}            & 69.31  & 42.19          & 84.88 & 77.29 & 14.72 \\
TransUNet \cite{5_chen2021transunet}          & 130.22 & 102.00         & 80.16 & 71.32 & 15.32 \\
UNet \cite{4_ronneberger2015u}                & 31.78  & 374.34         & 84.04 & 76.23 & 13.31 \\
Mask R-CNN \cite{he2017mask}                  & 90.44  & 67.13          & 83.97 & 76.42 & 12.89 \\
ResNet101-UNet \cite{diakogiannis2020resunet} & 267.62 & 112.79         & 84.92 & 77.73 & 13.03 \\
U2Seg \cite{niu2024unsupervised}              & 86.47  & 88.46          & 75.85 & 70.13 & 16.07 \\
SegNet \cite{badrinarayanan2017segnet}        & 69.21  & 42.19          & 85.32 & 78.11 & 12.71 \\
FUSegNet \cite{dhar2024fusegnet}              & 67.22  & 25.82 & 86.43 & 79.41 & 16.06 \\
MambaVision-UNet \cite{hatamizadeh2025mambavision} & 35.22 & 21.90 & 85.25 & 77.91 & 13.19 \\
MedMamba-UNet \cite{yue2024medmamba}          & \textbf{18.36} & 43.30    & 84.16 & 76.30 & 15.74 \\
LSS-Mamba                                     & 35.39  & 46.85          & 85.96 & 78.68 & 16.15 \\
\hline
\rowcolor{gray!20}
\textbf{LINDA (Ours)} & 25.70 & 82.45 & \textbf{86.68} & 79.67 & \textbf{6.37} \\
\hline
\end{tabular}
\\[2pt]

\end{table}

\subsection{Comparison with State-of-the-art Methods}
Table~\ref{tab:sota_comparison} compares LINDA with segmentation models re-trained under one protocol, which keeps the comparison fair. Under this setting LINDA gives the best Dice (86.68\%), IoU (79.67\%), and boundary accuracy (HD95 6.37 pixels) in the table. The boundary margin is the clearest: at 6.37 pixels the HD95 is roughly half that of the next-best model, which points to tighter agreement along the wound rim. LINDA reaches these numbers with 25.70\,M parameters, far fewer than heavy encoders such as ResNet101-UNet \cite{diakogiannis2020resunet}, TransUNet \cite{5_chen2021transunet}, and Mask R-CNN \cite{he2017mask}, which supports its use where computing is limited. FUSegNet \cite{dhar2024fusegnet}, the FUSeg 2021 challenge winner, is included as a domain-specific reference; under the same 100-epoch schedule it trails LINDA on all three metrics, while its originally published Dice, obtained with heavier training, is higher.

\begin{table}[!ht]
\centering
\caption{Ablation study progressively adding LSS, LTC, and BAL to the baseline.}
\label{tab:ablation}
\renewcommand{\arraystretch}{1.4}
\begin{tabular}{ >{\arraybackslash}m{4.7cm} | >{\centering\arraybackslash}m{0.9cm} >{\centering\arraybackslash}m{0.9cm} >{\centering\arraybackslash}m{0.9cm} | >{\centering\arraybackslash}m{1.3cm} >{\centering\arraybackslash}m{1.3cm} >{\centering\arraybackslash}m{1.3cm} }
\hline
\textbf{Model Variation} & \textbf{LSS} & \textbf{LTC} & \textbf{BAL} & \textbf{Dice \newline (\%)($\uparrow$)} & \textbf{IoU \newline (\%)($\uparrow$)} & \textbf{Params \newline (M)} \\
\hline
Baseline (ResNet-34 + U-Net) & & & & $85.22$ & $77.56$ & 24.87 \\
Baseline + LSS & \checkmark & & & $83.51$ & $75.51$ & 25.12 \\
Baseline + LTC & & \checkmark & & $83.29$ & $75.22$ & 25.45 \\
Baseline + LSS + BAL & \checkmark & & \checkmark & $86.41$ & $79.26$ & 25.12 \\
LSS + Decoder only + BAL & \checkmark & & \checkmark & $46.09$ & $34.61$ & 3.01 \\
LINDA (w/o BAL) & \checkmark & \checkmark & & $76.18$ & $69.74$ & 25.70 \\
\hline
\rowcolor{gray!20}
\textbf{LINDA (Proposed)} & \checkmark & \checkmark & \checkmark & $\mathbf{86.68}$ & $\mathbf{79.67}$ & $\mathbf{25.70}$ \\
\hline
\end{tabular}
\end{table}
\vspace{-0.3cm}
\subsection{Ablation Studies}
\textbf{Model Components:} Table~\ref{tab:ablation} reports the ablation on different components of the proposed model. Adding LSS and LTC to the ResNet-34 baseline without the Boundary Alignment Loss (BAL) drops Dice from 85.22\% to 76.18\%: pixel-wise losses alone do not bring the structural prior and the refinement together. BAL fixes this. With LSS and BAL the model reaches 86.41\% Dice, and the LTC loop takes the full model to 86.68\% Dice and 79.67\% IoU. Two further rows isolate the rest: keeping only LSS with the decoder, without the pretrained encoder, collapses to 46.09\% Dice, so the encoder is essential, while the gap from LSS+BAL to the full model is the LTC contribution. The full model adds only 0.83\,M parameters over the baseline, and this pattern reinforces the interpretability point: the LSS prior is used well only once the boundary loss ties the prediction to it.

\begin{table}[t]
\centering
\caption{Effect of the number of LTC refinement steps $T$ on the FUSeg validation set. Dice/IoU in \%, HD95 in pixels; $T{=}4$ (shaded) is used.}
\label{tab:tsteps}
\renewcommand{\arraystretch}{1.2}
\setlength{\tabcolsep}{5pt}
\begin{tabular}{ c | c c c || c | c c c }
\hline
\textbf{T} & \textbf{Dice} ($\uparrow$) & \textbf{IoU} ($\uparrow$) & \textbf{HD95} ($\downarrow$) & \textbf{T} & \textbf{Dice} ($\uparrow$) & \textbf{IoU} ($\uparrow$) & \textbf{HD95} ($\downarrow$) \\
\hline
1 & 86.28 & 79.33 & 6.93  & 8  & 86.26 & 79.33 & 6.64  \\
2 & 85.27 & 78.10 & 11.36 & 9  & 69.21 & 57.54 & 30.91 \\
3 & 84.66 & 77.12 & 9.71  & 10 & 67.22 & 55.67 & 31.02 \\
\cellcolor{gray!20}4 & \cellcolor{gray!20}86.68 & \cellcolor{gray!20}79.67 & \cellcolor{gray!20}6.37 & 11 & 72.47 & 62.35 & 22.92 \\
5 & 85.55 & 78.31 & 9.34  & 16 & 69.79 & 58.55 & 27.23 \\
6 & 86.65 & 79.57 & 6.59  & 20 & 48.72 & 41.26 & 48.35 \\
7 & 86.21 & 79.11 & 6.33  & 32 & 68.57 & 57.62 & 30.63 \\
\hline
\end{tabular}
\end{table}

\textbf{Number of refinement steps:} Table~\ref{tab:tsteps} varies the number of LTC steps $T$. Performance is stable for $T{=}1$ to $T{=}8$, and $T{=}4$ gives the best balance of overlap (86.68\% Dice, 79.67\% IoU) and boundary accuracy (HD95 6.37) at a low step count. Beyond $T{=}8$ it collapses, with Dice below 73\% and HD95 above 22 pixels, consistent with the Euler-discretized ODE becoming unstable over too many steps. We therefore use $T{=}4$.

\vspace{-0.3cm}
\subsection{A Closer Look at the LSS Maps}
\begin{figure}[!htb]
\centering
\includegraphics[width=11cm, height=5.1cm]{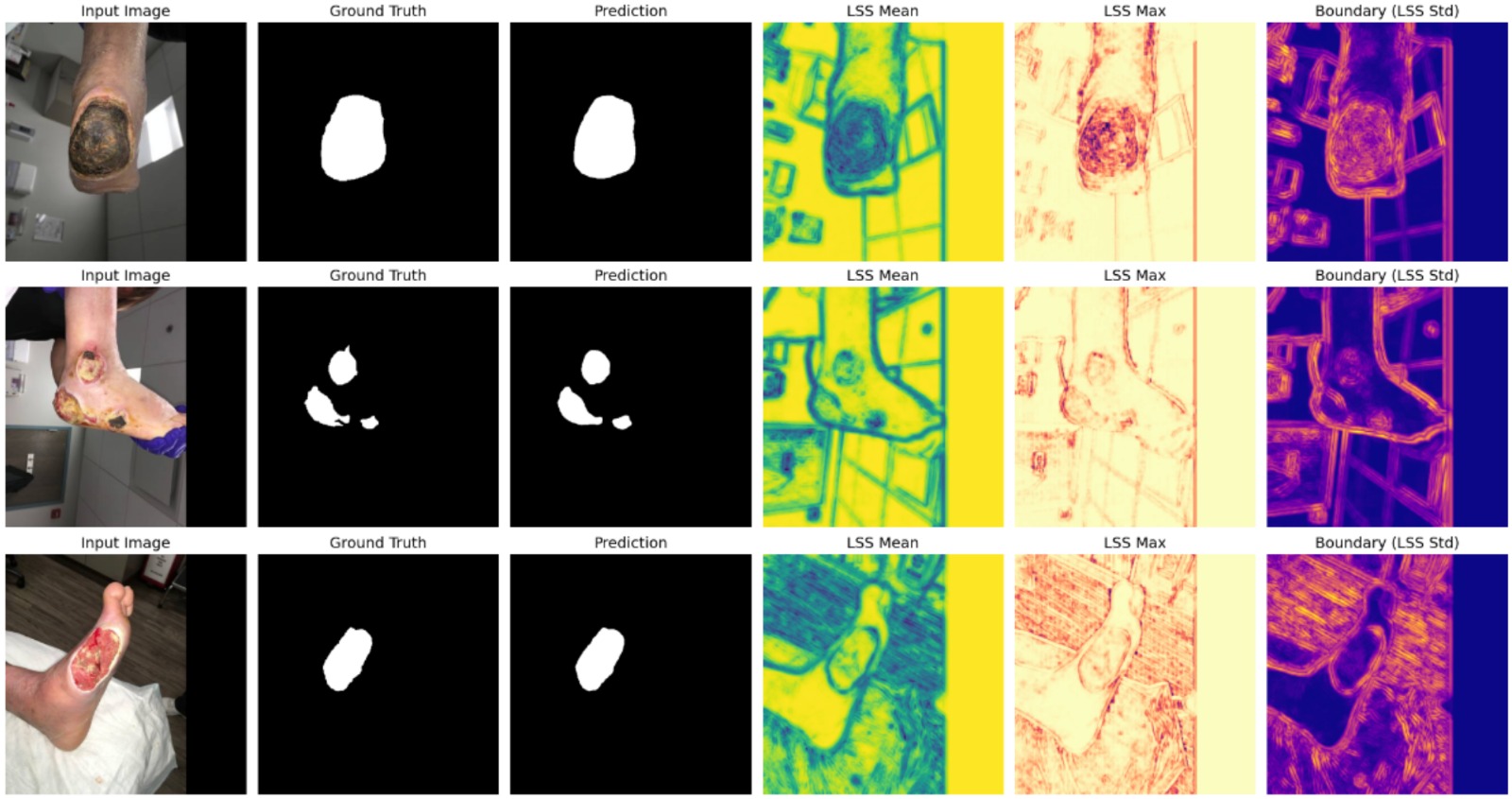}
\caption{The three LSS channels (Mean, Max, and Std) for sample wounds. The Std channel highlights the wound rim, while Mean and Max capture local homogeneity and strong local structure.}
\label{fig3}
\end{figure}

Figure~\ref{fig3} shows the three LSS channels for several wounds. The Std channel behaves like a local edge map: it stays quiet over uniform background and responds along the wound rim, which is exactly where boundary errors are most costly. These maps feed the encoder, and the Mean channel is the target of the boundary loss (Eq.~\ref{eq:bal_loss}), so they give a direct, image-level account of the structural information the model is pushed to respect. We therefore treat them as a way to inspect the model's inputs and training signal, not as a measurement of how the deeper layers reach their final decision.

\vspace{-1cm}
\section{Discussion}
We presented \textbf{LINDA}, a foot ulcer segmentation model that fuses a deterministic Local Self-Similarity prior into a ResNet-34 encoder and refines the prediction with a small Liquid Time-Constant module and a boundary alignment loss. It produces accurate masks with the tightest boundaries in our comparison at only $25.70$\,M parameters, and since the LSS maps are computed before encoding, they also give a readable view of the cues the model follows, with no separate explanation step, thus making it a faithful and explainable segmentation model. 
\vspace{-0.4cm}
\section{Impact in RCS}
In resource constrained setting, it is crucial to balance performance, model size and inference speed. Our proposed approach, achieves the best tradeoff between model size ($25.70$\,M parameters), training time ($\sim5$ hrs) and per-image inference time ($\sim45$ ms). LINDA achieves state-of-the-art scores in Dice and HD95 metrics with a reasonably low number of parameters and GFLOPS, as compared to its closest competitors, thus making it an excellent choice for a wound segmentation model in a resource constrained setting.

\begin{credits}
\subsubsection{\ackname} No acknowledgments or competing interests.

\subsubsection{\discintname}
The authors have no competing interests to declare that are
relevant to the content of this article.
\end{credits}

\bibliographystyle{unsrt}
\bibliography{references}

@article{1_armstrong2017diabetic,
  title={Diabetic foot ulcers and their recurrence},
  author={Armstrong, David G and Boulton, Andrew JM and Bus, Sicco A},
  journal={New England Journal of Medicine},
  volume={376},
  number={24},
  pages={2367--2375},
  year={2017},
  publisher={Mass Medical Soc}
}

@article{2_cassidy2021dfuc2020,
  title={The DFUC 2020 dataset: Analysis towards diabetic foot ulcer detection},
  author={Cassidy, Bill and Reeves, Neil D and Pappachan, Joseph M and Gillespie, David and O'Shea, Claire and Rajbhandari, Sanjeev and Maiya, Arun G and Frank, Eibe and Boulton, Andrew JM and Yap, Moi Hoon},
  journal={IEEE Journal of Biomedical and Health Informatics},
  volume={26},
  number={1},
  pages={346--356},
  year={2021},
  publisher={IEEE}
}

@article{3_goyal2020recognition,
  title={Recognition of ischemia and infection in diabetic foot ulcers: Dataset and techniques},
  author={Goyal, Manu and Reeves, Neil D and Davison, Andrew K and Rajbhandari, Sanjeev and Sprague, John and Yap, Moi Hoon},
  journal={Computers in Biology and Medicine},
  volume={117},
  pages={103616},
  year={2020},
  publisher={Elsevier}
}

@inproceedings{4_ronneberger2015u,
  title={U-net: Convolutional networks for biomedical image segmentation},
  author={Ronneberger, Olaf and Fischer, Philipp and Brox, Thomas},
  booktitle={Medical Image Computing and Computer-Assisted Intervention--MICCAI 2015},
  pages={234--241},
  year={2015},
  organization={Springer}
}

@inproceedings{hatamizadeh2025mambavision,
  title     = {{MambaVision}: A Hybrid Mamba-Transformer Vision Backbone},
  author    = {Hatamizadeh, Ali and Kautz, Jan},
  booktitle = {Proceedings of the IEEE/CVF Conference on Computer Vision and Pattern Recognition (CVPR)},
  year      = {2025}
}

@article{yue2024medmamba,
  title   = {{MedMamba}: Vision Mamba for Medical Image Classification},
  author  = {Yue, Yubiao and Li, Zhenzhang},
  journal = {arXiv preprint arXiv:2403.03849},
  year    = {2024}
}

@article{dhar2024fusegnet,
  title     = {{FUSegNet}: A deep convolutional neural network for foot ulcer segmentation},
  author    = {Dhar, Mrinal Kanti and Zhang, Taiyu and Patel, Yash and Gopalakrishnan, Sandeep and Yu, Zeyun},
  journal   = {Biomedical Signal Processing and Control},
  volume    = {92},
  pages     = {106057},
  year      = {2024},
  publisher = {Elsevier}
}

@article{5_chen2021transunet,
  title={Transunet: Transformers make strong encoders for medical image segmentation},
  author={Chen, Jieneng and Lu, Yongyi and Yu, Qihang and Luo, Xiangde and Adeli, Ehsan and Wang, Yan and Lu, Le and Yuille, Alan L and Zhou, Yuyin},
  journal={arXiv preprint arXiv:2102.04306},
  year={2021}
}

@article{6_hatamizadeh2022unetr,
  title={Unetr: Transformers for 3d medical image segmentation},
  author={Hatamizadeh, Ali and Tang, Yucheng and Nath, Vishwesh and Tseng, Dong and Idrees, Juan and Li, Anwar and Roth, Holger and Xu, Daguang},
  journal={Proceedings of the IEEE/CVF Winter Conference on Applications of Computer Vision},
  pages={574--584},
  year={2022}
}

@inproceedings{7_hasani2021liquid,
  title={Liquid time-constant networks},
  author={Hasani, Ramin and Lechner, Mathias and Amini, Alexander and Rus, Daniela and Grosu, Radu},
  booktitle={Proceedings of the AAAI Conference on Artificial Intelligence},
  volume={35},
  number={9},
  pages={7657--7666},
  year={2021}
}

@ARTICLE{11205342,
  author={Chakrabarty, Sudip and Bishwas, Pappu and Bandyopadhyay, Mainak and Sublime, Jérémie},
  journal={IEEE Access}, 
  title={Can We Trust AI With Our Ears? A Cross-Domain Comparative Analysis of Explainability in Audio Intelligence}, 
  year={2025},
  volume={13},
  number={},
  pages={179733-179758},
  doi={10.1109/ACCESS.2025.3622161}}

@article{akman2024audio,
  title={Audio explainable artificial intelligence: A review},
  author={Akman, Alican and Schuller, Bj{\"o}rn W},
  journal={Intelligent Computing},
  volume={3},
  pages={0074},
  year={2024},
  publisher={AAAS}
}

@article{van2022explainable,
  title={Explainable artificial intelligence (XAI) in deep learning-based medical image analysis},
  author={Van der Velden, Bas HM and Kuijf, Hugo J and Gilhuijs, Kenneth GA and Viergever, Max A},
  journal={Medical image analysis},
  volume={79},
  pages={102470},
  year={2022},
  publisher={Elsevier}
}

@article{badrinarayanan2017segnet,
  title={Segnet: A deep convolutional encoder-decoder architecture for image segmentation},
  author={Badrinarayanan, Vijay and Kendall, Alex and Cipolla, Roberto},
  journal={IEEE TPAMI},
  volume={39},
  number={12},
  pages={2481--2495},
  year={2017}
}

@inproceedings{diakogiannis2020resunet,
  title={Resunet-a: A deep learning framework for semantic segmentation of remote sensing data},
  author={Diakogiannis, Foivos I and Waldner, Fran{\c{c}}ois and Caccetta, Peter and Wu, Chen},
  journal={ISPRS Journal of Photogrammetry and Remote Sensing},
  volume={162},
  pages={94--114},
  year={2020}
}

@article{wang2024fuseg,
  title={FUSeg: The foot ulcer segmentation challenge},
  author={Wang, Chuanbo and Mahbod, Amirreza and Ellinger, Isabella and Galdran, Adrian and Gopalakrishnan, Sandeep and Niezgoda, Jeffrey and Yu, Zeyun},
  journal={Information},
  volume={15},
  number={3},
  pages={140},
  year={2024},
  publisher={MDPI}
}

@inproceedings{alkhalefah2025advancing,
  title={Advancing diabetic foot ulcer care: AI and generative AI approaches for classification, prediction, segmentation, and detection},
  author={Alkhalefah, Suhaylah and AlTuraiki, Isra and Altwaijry, Najwa},
  booktitle={Healthcare},
  volume={13},
  number={6},
  pages={648},
  year={2025},
  organization={MDPI}
}

@article{bhalla2023discriminative,
  title={Discriminative feature attributions: Bridging post hoc explainability and inherent interpretability},
  author={Bhalla, Usha and Srinivas, Suraj and Lakkaraju, Himabindu},
  journal={Advances in neural information processing systems},
  volume={36},
  pages={44105--44122},
  year={2023}
}

@inproceedings{simonyan2015very,
  title={Very deep convolutional networks for large-scale image recognition},
  author={Simonyan, Karen and Zisserman, Andrew},
  booktitle={International Conference on Learning Representations (ICLR)},
  year={2015}
}

@inproceedings{he2017mask,
  title={Mask {R-CNN}},
  author={He, Kaiming and Gkioxari, Georgia and Doll{\'a}r, Piotr and Girshick, Ross},
  booktitle={Proceedings of the IEEE International Conference on Computer Vision (ICCV)},
  pages={2961--2969},
  year={2017}
}

@inproceedings{niu2024unsupervised,
  title={Unsupervised universal image segmentation},
  author={Niu, Dantong and Wang, Xudong and Han, Xinyang and Lian, Long and Herzig, Roei and Darrell, Trevor},
  booktitle={Proceedings of the IEEE/CVF Conference on Computer Vision and Pattern Recognition (CVPR)},
  year={2024}
}

@inproceedings{shechtman2007matching,
  title={Matching local self-similarities across images and videos},
  author={Shechtman, Eli and Irani, Michal},
  booktitle={IEEE Conference on Computer Vision and Pattern Recognition (CVPR)},
  pages={1--8},
  year={2007},
  organization={IEEE}
}

@inproceedings{oota2023wsnet,
  title={{WSNet}: Towards an effective method for wound image segmentation},
  author={Oota, Subba Reddy and Rowtula, Vijay and Mohammed, Shahid and Liu, Minghu and Gupta, Manish},
  booktitle={Proceedings of the IEEE/CVF Winter Conference on Applications of Computer Vision (WACV)},
  pages={3234--3243},
  year={2023}
}

@article{wang2020fully,
  title={Fully automatic wound segmentation with deep convolutional neural networks},
  author={Wang, Chuanbo and Anisuzzaman, D. M. and Williamson, Victor and Dhar, Mrinal Kanti and Rostami, Behrouz and Niezgoda, Jeffrey and Gopalakrishnan, Sandeep and Yu, Zeyun},
  journal={Scientific Reports},
  volume={10},
  number={1},
  pages={21897},
  year={2020},
  publisher={Nature Publishing Group}
}

@article{nataliani2025dfu,
  title={{DFU-MambaLiteUNet}: A lightweight and efficient model for diabetic foot ulcer segmentation},
  author={Nataliani, Yessica and Purnomo, Hindriyanto Dwi and Timotius, Ivanna Kristianti and Purwono, Purwono},
  journal={Expert Systems with Applications},
  volume={293},
  pages={128689},
  year={2025},
  publisher={Elsevier}
}

@inproceedings{borst2025woundambit,
  title={{WoundAmbit}: Bridging state-of-the-art semantic segmentation and real-world wound care},
  author={Borst, Vanessa and Dittus, Timo and Dege, Tassilo and Schmieder, Astrid and Kounev, Samuel},
  booktitle={Machine Learning and Knowledge Discovery in Databases. Applied Data Science Track (ECML PKDD)},
  year={2025},
  organization={Springer}
}

@misc{thomas2017medetec,
  title={Medetec wound database},
  author={Thomas, Stephen},
  year={2017},
  note={http://www.medetec.co.uk}
}
\end{document}